\pdfoutput=1
\documentclass[10pt,twocolumn,letterpaper]{article}
\usepackage[T1]{fontenc}
\usepackage[pagenumbers]{cvpr}
\usepackage{graphicx,amsmath,amssymb,booktabs,tabularx,multirow}
\usepackage{capt-of}
\usepackage{xcolor}
\definecolor{cvprblue}{rgb}{0.21,0.49,0.74}
\usepackage[pagebackref,breaklinks,colorlinks,allcolors=cvprblue]{hyperref}
\def\confName{CVPR}
\def\confYear{2027}
\newcommand{\method}{CineAct}
\title{CineAct: Interaction-State Agents for Playable, Branching Film Generation}
\author{
Junyan Lin\textsuperscript{1,7,*} \quad
Jianqi Ma\textsuperscript{2,1,*,\ddag} \quad
Shujie Duan\textsuperscript{3} \quad
Hanxin Zhu\textsuperscript{4} \quad
Zhihao Hu\textsuperscript{2} \quad
Jinming Liu\textsuperscript{5} \\
Shuai Li\textsuperscript{6,1} \quad
Jun Zhang\textsuperscript{2,\dag} \quad
Xin Jin\textsuperscript{7,\dag} \quad
Chang Wen Chen\textsuperscript{1,\dag} \quad
Weidong Zhang\textsuperscript{2} \\[4pt]
{\small \textsuperscript{1}Hong Kong Polytechnic University \quad \textsuperscript{2}Tencent AIPD \quad \textsuperscript{3}Xiamen University Malaysia \quad \textsuperscript{4}University of Science and Technology of China} \\
{\small \textsuperscript{5}Shanghai Jiao Tong University \quad \textsuperscript{6}Northwest Polytechnical University \quad \textsuperscript{7}Eastern Institute of Technology, Ningbo} \\[2pt]
{\footnotesize \textsuperscript{*}Equal contribution. \quad \textsuperscript{\ddag}Project leader. \quad \textsuperscript{\dag}Corresponding authors.}
}
\begin{document}
\twocolumn[{\renewcommand\twocolumn[1][]{#1}
\maketitle
\begin{center}
\centering\includegraphics[width=\textwidth]{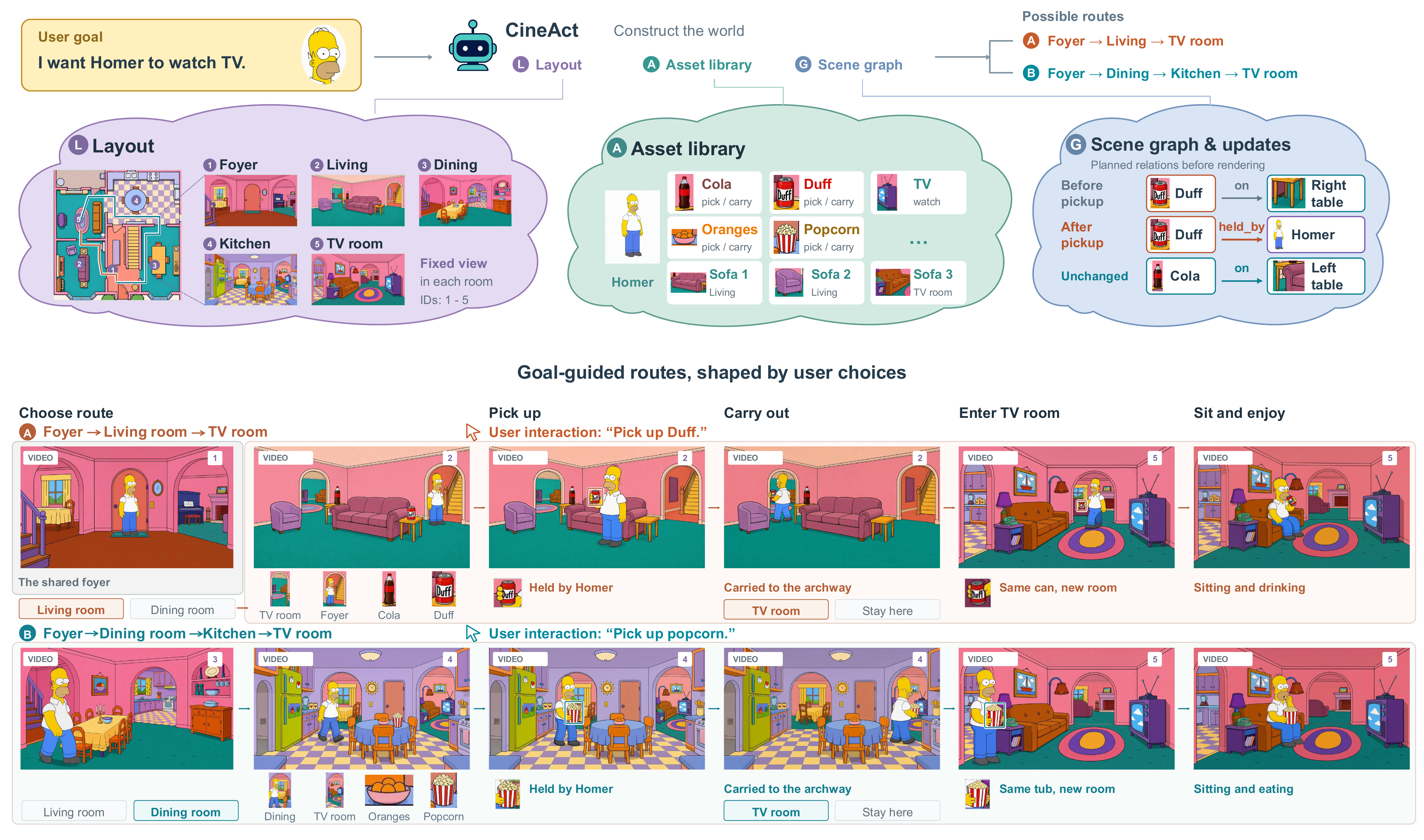}
\captionof{figure}{\textbf{Goal-grounded branching with \method{}.} Given a shared goal (\emph{``I want to watch TV''}), \method{} initializes a spatial layout ($L$), an asset library ($\mathcal{A}$), and a scene graph ($G$) to govern route selection, entity identity, and state changes at each turn. Whether Homer takes Duff from the living room (Route~A) or popcorn from the kitchen (Route~B), the chosen approach is executed on screen and persists across room cuts until he sits in the TV room.}
\label{fig:teaser}
\end{center}
}]
\begin{abstract}
Open video generators such as MiniMax-H3 produce short clips of only several to a dozen seconds per run. To synthesize longer sequences, existing methods either rely on agentic systems following preset scripts, or roll out autoregressively from the previous clip's last frame, losing off-screen states and accumulating hallucinations across turns. Consequently, users still cannot take an explicit goal, choose how to accomplish it, and see that choice persist across subsequent clips. We study goal-grounded interactive video generation, where one goal admits multiple valid approaches, the user's chosen approach must be executed on screen, and its consequences must persist across clips. To this end, we propose CineAct, a state-first agentic framework in which language-model agents handle reasoning while frozen image and video generators serve purely as renderers. Before interaction, the Architect initializes the world as a spatial layout with cameras, an asset library, and a scene graph; at each turn, the system deduces the planned state change, generates a target end frame with an image generator and rigorously verifies it, and synthesizes intermediate motion with an FL2V model. We further introduce a multi-space benchmark covering two suites: long-horizon step execution and open-intent branching admissibility. Experiments show that explicit world-state modeling boosts multi-step rollout consistency over the pixel-only baseline (+11.4 automated, +23.1 human points) and raises paired branching success toward a shared goal by +30.8 points.
\end{abstract}

\section{Introduction}
Open video generators such as Wan~\cite{wan}, HunyuanVideo~\cite{hunyuanvideo}, CogVideoX~\cite{cogvideox}, and Open-Sora~\cite{opensora}, alongside models such as MiniMax-H3~\cite{minimaxh3}, have made substantial progress in visual quality and motion realism, yet a single generation remains confined to short clips of several to a dozen seconds. To go beyond this limit and assemble complete long-form narratives, agentic director systems such as MovieAgent~\cite{movieagent}, ViMax~\cite{vimax}, and MVAgent~\cite{mvagent}, together with node-based multimodal creative canvases, chain high-level scripts or keyframes into continuous multi-shot videos. In parallel, interactive video models such as Genie~\cite{genie,genieredux}, GameNGen~\cite{gamengen}, and WorldPlay~\cite{worldplay} allow users to navigate through generated environments in real time via reactive controls. Recent work further explores combining the two lines: an agent plans the next action from a task goal, a video model renders it, and the system checks whether the action was completed on screen~\cite{collabvr,worldguide,oneira}.

Yet how users interact with these systems, and how many routes they can take, remains tightly restricted. Agentic directors and creative canvases organize shots around a preset script; their review and resampling~\cite{vimax,mvagent} enforce a single storyboard, and turning them into interactive interfaces merely splits a fixed workflow into sequential clicks. Interactive video models support real-time camera navigation~\cite{matrixgame,worldplay}, instruction following~\cite{gamecraft2,incantation}, or object-level trajectory and action persistence~\cite{worldcraft,actworld}, yet they execute step-level controls without offering alternative task-level routes toward a goal. Conversely, goal-driven planning systems advance toward a task objective~\cite{collabvr,worldguide,oneira}, but the planner autonomously determines the action sequence rather than letting the user choose among distinct valid approaches. In many domains, however, choosing the approach is the core interaction: trying alternative tools in skill training, valuing solution diversity as a key metric of engagement in playable games, testing task robustness across approaches in embodied evaluation, and leaving lasting traces in interactive narratives. Multiple valid paths toward a shared goal also provide an environment for agents to explore, compare action consequences, and accumulate verifiable experience toward policy learning and recursive self-improvement~\cite{voyager,stop,rise,agentgarten}. We therefore study \emph{goal-grounded interactive video generation}, supporting user-chosen routes toward a shared goal via menus, clicks, drags, or text, executing them visibly on screen, and preserving their consequences across turns (Fig.~\ref{fig:teaser}).

While language agents can readily propose diverse alternative branches, existing video generators lack the physical grounding to execute them reliably. The core bottleneck is that current pipelines roll out autoregressively from the previous clip's last frame alone, leaving the entire world state implicit in pixels. Relying solely on pixels creates three blind spots: the system lacks explicit preconditions to determine which actions are feasible and keep the goal reachable, cannot preserve object identities and states once entities leave the view or look-alike distractors appear, and has no geometric reference to align doorway directions and camera cuts across spaces. Thus, minor hallucinations compound over turns as removed items reappear, states revert on revisits, and characters take the wrong door, erasing distinctions between user choices.

To overcome these blind spots, we propose \method{}, a state-first agentic framework decoupling high-level world evolution from low-level pixel synthesis (Fig.~\ref{fig:framework}): an explicit world state governs \emph{what exists and changes}, while frozen image and video models render \emph{how it looks and moves}. Addressing the three challenges above, \method{} maintains a spatial layout with cameras, an asset library, and a scene graph to constrain cross-space transitions, visual identity, and goal-reaching feasibility. At each turn, agents ground user input on this state to deduce a transition specification, synthesize and verify a target end frame to lock the endpoint, and condition an FL2V model under camera-geometry rules to render motion whose last frame refreshes $\mathcal{A}_t$ and seeds the next turn.

To evaluate this task systematically, we introduce a multi-space benchmark covering long-horizon step execution and branching admissibility. Across action, world-state, and spatial dimensions, we conduct controlled ablations, cross-backbone comparisons, counterfactual gating probes, paired branching tests, and an interface study. In multi-step rollouts, \method{} substantially improves overall consistency over the pixel-only baseline across both automated (+11.4 points) and blind human ratings (+23.1 points), with ablations confirming distinct gains from layout, assets, and scene graphs. Moreover, paired branching evaluations confirm that \method{} gates, executes, and preserves diverse user-chosen routes toward a shared goal (+30.8 points in paired branching success over Baseline).

Our contributions are threefold:
\begin{itemize}
\item \textbf{Task formulation.} We formalize goal-grounded interactive video generation, requiring the system to support multiple user-chosen routes toward a shared goal while jointly satisfying action validity, execution visibility, and cross-turn consequence persistence.
\item \textbf{State-first agentic framework.} We propose \method{}, which decouples world-state reasoning from pixel synthesis. By maintaining explicit spatial layouts, dynamically refreshed asset libraries, and scene graphs, \method{} locks in post-action endpoints via pre-rendering target-frame verification and drives FL2V generation under camera-geometry rules.
\item \textbf{Benchmark and evaluation.} We introduce a multi-space benchmark covering long-horizon step execution and branching admissibility across action, world-state, and spatial dimensions. Controlled ablations, backbone studies, and interface comparisons quantify each representation's distinct contribution.
\end{itemize}

\section{Related Work}
\paragraph{Multi-shot video and agentic directors.}
Agentic director systems orchestrate multiple shots to synthesize coherent long-form videos from narrative scripts. Some simulate scriptwriting, storyboarding, and shooting via hierarchical chain-of-thought~\cite{movieagent} or continuously monitor and refine narrative coherence with vision--language agents~\cite{vimax}. MVAgent~\cite{mvagent} maintains a continuity memory and samples candidate continuations from a shared shot state; CineEvolve~\cite{cineevolve} co-evolves director and critic policies; and VideoGen-Agent~\cite{videogen} coordinates generation and verification tools through reinforcement learning. To support non-consecutive revisits to prior spaces, ATAS~\cite{atas} compiles screenplays into bounded shot requests backed by explicit scene-geometry and event memories. Although these systems produce coherent multi-shot films, they remain organized around preset scripts rather than interactive user choices toward a task goal. We instead require every generated clip to respond directly to the user's online choice across multiple viable routes.

\paragraph{Controllable and interactive video generation.}
Interactive video models roll out frames in pixel space in response to control signals or online user inputs. Early work synthesizes interactive environments via unsupervised latent actions~\cite{genie}, exploration-driven data collection~\cite{genieredux}, or agent replay~\cite{gamengen}. Subsequent streaming models enable frame-level keyboard-and-mouse control~\cite{matrixgame}, maintain long-range geometric consistency by dynamically rebuilding context memory from past frames~\cite{worldplay}, or follow open-ended instruction prompts in game worlds~\cite{gamecraft2}. Recent work advances to entity-level manipulation: Incantation~\cite{incantation} introduces per-entity language control with task-specific status tracking, WorldCraft~\cite{worldcraft} combines object selection and 2D drag trajectories with camera navigation to preserve object placements across off-screen revisits, and ActWorld~\cite{actworld} retains action-induced states such as opened doors and picked-up items via action-aware memory. In parallel, goal-driven systems such as CollabVR~\cite{collabvr} and WorldGuide~\cite{worldguide} couple step-level action planning with video execution verification toward a task goal. However, these approaches either execute step-level controls without checking multi-step goal reachability, or rely on an automated planner to dictate the next step without offering users alternative, equally valid routes toward a shared goal.

\paragraph{Explicit world states and generative rendering.}
To curb hallucination accumulation in autoregressive rollouts, recent work decouples state logic and physical rules from neural pixel synthesis. PWM~\cite{pwm} compiles language into world programs and guides rendering with 3D bounding boxes and a trained spatial control branch; AgentGarten~\cite{agentgarten} constructs code-driven worlds for evolving agents; and GameDirector~\cite{gamedirector} and Code World Model~\cite{cwm} separate combat mechanics or executable code evolution from video rendering. For open-world interaction, Oneira~\cite{oneira} decomposes goals into dependency-ordered interactions over an object state table with LoRA-adapted rendering, while State World Models~\cite{stateworldmodels} combine a typed rule ledger and goal conditions with clay-scene camera rendering. Within single-clip synthesis, CausalMotion~\cite{causalmotion} and PhysPlan~\cite{physplan} steer inference using causal keyframes or sparse edits to a physical state graph, and SolarRenderer~\cite{solarrenderer} injects structural videos through the backbone's native reference pathway. While explicit state representations are well established, existing frameworks use them to execute predefined mechanics or a single decomposed plan, and several require training task-specific adapters or control branches~\cite{pwm,cwm,oneira,gamedirector}. We instead pair frozen image and video generators with a spatial layout, a refreshed asset library, and a scene graph to deduce and validate multiple user-chosen routes toward a shared goal, using pre-rendering target-frame verification to lock in endpoints before FL2V synthesis.

\section{Problem Setting}
\begin{figure*}[t]
\centering\includegraphics[width=0.98\textwidth]{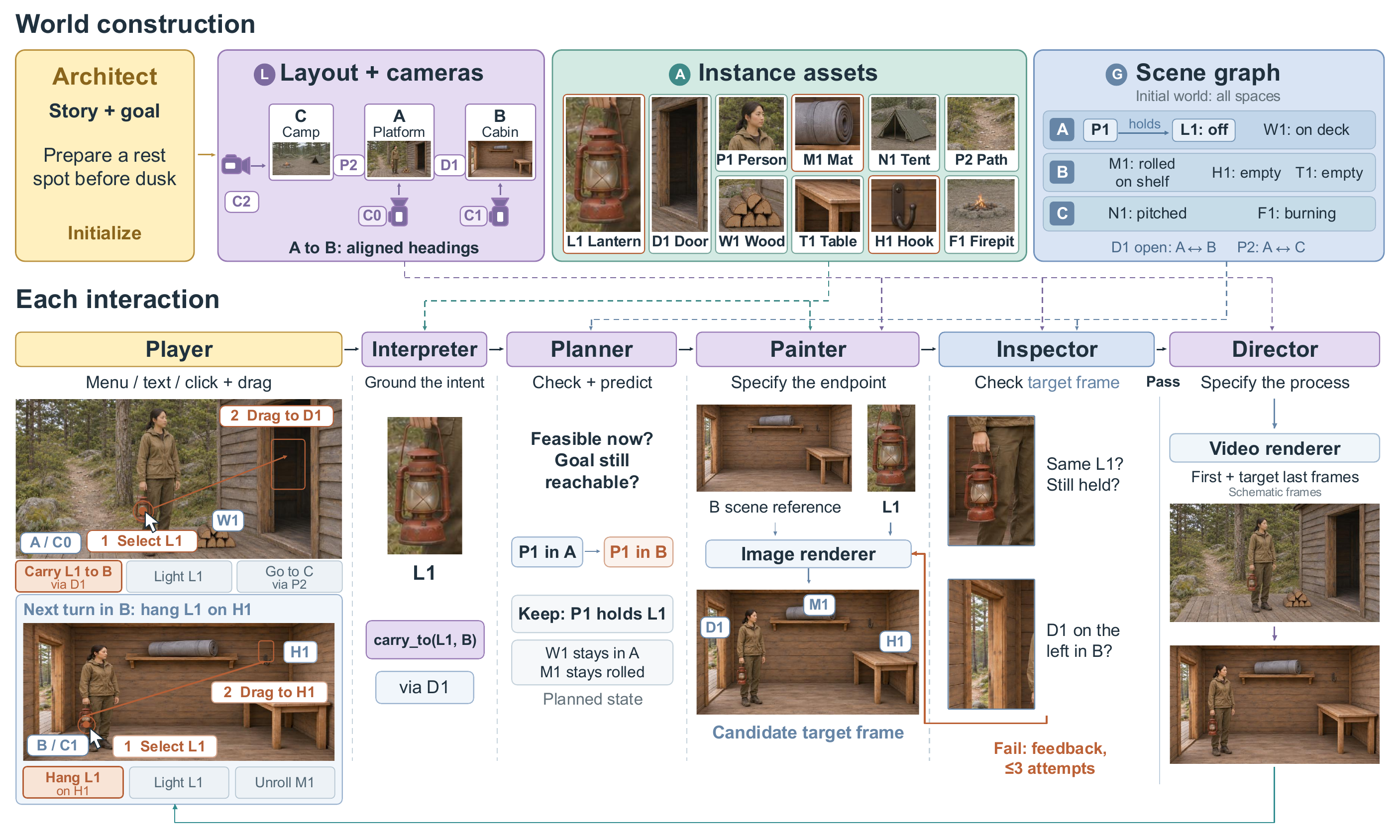}
\caption{\textbf{\method{} framework.} \textbf{Top (World construction):} From a story goal, Architect initializes a spatial layout with fixed cameras ($L$), an asset library of instance and scene references ($\mathcal{A}$), and a multi-space scene graph ($G$). \textbf{Bottom (Each interaction):} Interpreter grounds the player's input, Planner checks preconditions and goal reachability over $G$, Painter drives an image renderer to synthesize a candidate target end frame, and Inspector verifies it against the planned state (returning failures for up to three attempts) before Director renders the transition video whose tail refreshes $\mathcal{A}$ and seeds the next turn.}
\label{fig:framework}
\end{figure*}
We formulate goal-grounded interactive video generation as a multi-turn, branchable conditional generation problem organized around a shared task goal. Let $g$ be the global goal to achieve (e.g., ``prepare a rest spot before dusk'') starting from an initial scene observation $I_1$. At turn $t$, the user provides an interaction input $u_t \in \mathcal{U}$ via a menu selection, a click or drag trajectory on the frame, or free text. Conditioning on the current frame $I_t$ and interaction history, the system resolves $u_t$ into a semantic action $a_t$ and synthesizes a video clip $V_t$, whose actual last frame $I_{t+1} = \mathrm{Tail}(V_t)$ serves as the starting frame of turn $t+1$. Interaction continues until $g$ is accomplished, yielding a trajectory
\begin{equation}
  \tau = (I_1, u_1, V_1, I_2, u_2, V_2, \dots, u_K, V_K).
\end{equation}
Unlike single-script video generation, a shared starting state or intermediate history $(I_1, \dots, I_t)$ admits multiple valid routes $\{\tau^{(m)}\}$ that reach the same goal $g$ while leaving route-specific consequences in later frames.

\paragraph{Consistency requirements.}
For a generated trajectory $\tau$ to support interactive play, it must jointly satisfy three consistency criteria across turns $t \in \{1, \dots, K\}$:

\noindent\textbf{(1) Action validity and process visibility.}
Each executed action $a_t$ must satisfy the physical preconditions $P(a_t)$ of the current scene while keeping the global goal $g$ reachable in future turns. Moreover, the generated clip $V_t$ must visibly realize the physical action process rather than merely teleporting to the end state:
\begin{equation}
  V_t \models \Delta(a_t).
\end{equation}

\noindent\textbf{(2) Entity identity and state persistence.}
Manipulated objects and characters must preserve their visual identity without vanishing, duplicating, or swapping across turns. Both the newly established holding or support states and the unaffected scene attributes $K(a_t)$ must persist across subsequent clips (when entities leave and re-enter view):
\begin{equation}
  V_k \models K(a_t), \quad \forall\, k \ge t.
\end{equation}

\noindent\textbf{(3) Cross-space geometric and camera continuity.}
When interactions span multiple connected spaces or revisit prior spaces, the character's exit direction from the source space, entry portal side in the destination space, and camera viewpoint transitions must remain geometrically coherent across clips.
Standard autoregressive rollouts condition only on the previous frame $I_t$, leaving all three constraints implicit in pixels. Below, we introduce \method{} to explicitly model and enforce these constraints using frozen renderers.

\section{\method{}: A State-First Agentic Framework for Interactive Film Generation}
Fig.~\ref{fig:framework} gives an overview of \method{}. The framework decouples high-level reasoning from low-level rendering. Reasoning tasks, namely intent parsing, validity checking, reachability search, and target-frame verification, are handled by language-model agents. Image and video generators act purely as \emph{physical renderers}~\cite{solarrenderer,pwm}: they produce pixels under verified endpoint states and geometric constraints and take no part in deducing logical world state.

\subsection{Architect: World Initialization}
Before interaction begins, the language-model agent \textbf{Architect} constructs an interactive multi-space world around the global goal $g$ and designs multiple valid routes (e.g., Fig.~\ref{fig:framework}, top: carrying lantern L1 into cabin B or firewood W1 to camp C). To guarantee multi-route playability, the system verifies reachability during initialization so that at least two substantively distinct action sequences, differing in tools, strategy, or spatial path, can reach $g$ from the start or shared mid-story decision points without redundant round trips. Architect then materializes this world into three complementary representations $S_0 = (L, \mathcal{A}_0, G_0)$ (Fig.~\ref{fig:framework}, top):

\noindent\textbf{(1) Spatial layout and cameras ($L = (\mathcal{R}, \mathcal{P}, \mathcal{C})$).}
The layout represents the world as a 2D axis-aligned floor plan specifying rectangular space bounds $r \in \mathcal{R}$ and shared-boundary portal segments $D \in \mathcal{P}$, together with one deterministic fixed camera $\mathcal{C}$ per space (placed on a portal-free wall facing an exit). Geometric projection derives each portal's on-screen side (left, right, or far wall) to constrain cross-space transitions and verify doorway placement.

\noindent\textbf{(2) Asset library ($\mathcal{A}_t$).}
The asset library combines per-space visual memory (initialized from the opening renders and updated after each turn with the latest generated video tail of the visited space) with initial entity identity crops $\mathcal{A}_0(v)$ paired with unique identifiers and stable, location-free appearance descriptions. At each step, the system retrieves the destination-space memory and identity crops for action participants and off-screen actors while skipping characters already visible in $I_t$ to prevent duplicate figures.

\noindent\textbf{(3) Scene graph ($G_t = (\mathcal{V}_t, \mathcal{E}_t)$).}
Initialized as $G_0$, the scene graph tracks all in-view and off-screen entities $\mathcal{V}_t$. Containment, support, and holding relations are deterministically derived from hierarchical ownership pointers alongside portability, carrying-capacity limits, and state attributes $\mathcal{E}_t$ (e.g., open/closed, lit/unlit). A symbolic transition engine checks preconditions and goal reachability over $G_t$, supplying valid action suggestions at each turn.

\subsection{Interpreter and Planner: Intent Grounding and State Deduction}
At turn $t$, the system presents feasible menu suggestions derived from $G_t$, while the user may select a menu option or issue open-ended input $u_t$ via clicks, drags, or free text. Given the current frame $I_t$, the asset library $\mathcal{A}_t$, and $G_t$, the \textbf{Interpreter} grounds $u_t$ to specific scene entities, resolving it into a structured operation $a_t$, an expected state change $\Delta_t$, and invariance constraints (Fig.~\ref{fig:framework}, bottom).

The \textbf{Planner} then validates both in-plan choices and out-of-distribution (off-menu) operations using the symbolic transition engine over $G_t$. If $a_t$ matches a pre-verified branch, it advances directly; otherwise, the Planner checks physical preconditions $P(a_t)$, carrying capacity, and locality over $G_t$, deduces $G_t^{\mathrm{expected}} = G_t \oplus \Delta_t$, and runs graph search to verify that the global goal $g$ remains reachable from $G_t^{\mathrm{expected}}$. Inputs that are physically impossible, ambiguous, or would render $g$ unreachable are rejected before rendering. Validated actions advance the planned scene graph to $G_{t+1} = G_t^{\mathrm{expected}}$ (maintained separately from observed video frames) and deterministically diff $(G_t, G_{t+1})$ into an explicit \emph{transition specification}:
\begin{equation}
  \mathrm{Spec}_t = (\mathcal{V}_{\mathrm{operated}}, \Delta_t, \mathcal{V}_{\mathrm{kept}}, C_t),
\end{equation}
which details the operated instances, the modified relations, the instances that must remain untouched, and, for cross-space moves, the camera transition instruction $C_t$.

\subsection{Painter and Inspector: Verified Target-Frame Generation}
In first--last-frame (FL2V) video synthesis, the starting frame $I_t$ is fixed, so securing an accurate target end frame $I_t^\star$ provides reliable boundary anchors for rendering the motion in between. Given the transition specification, the \textbf{Painter} combines the planned changes, the layout $L$, and the asset library $\mathcal{A}_t$ into an image prompt and calls an image-editing model to generate a candidate target end frame $I_t^\star$ depicting the intended post-action state, conditioned on the current frame $I_t$, the destination-space memory on cross-space moves, and the operated object's identity crop (e.g., scene B and lantern L1 in Fig.~\ref{fig:framework}).

Between the Painter and the Director, the \textbf{Inspector} serves as a target-endpoint gatekeeper, using a vision--language model to verify each candidate target end frame item by item as pass, fail, or unclear before video rendering: whether it depicts the completed action, whether operated instances retain their identity, and whether non-target objects are unchanged; for cross-space moves, it also localizes doorways in the candidate frame and checks their screen sides against the geometric projections from $L$. Specifically, it verifies every entry of the transition specification, including the planned changes, held items, and must-keep instances (such as confirming in Fig.~\ref{fig:framework} that L1 remains the same held lantern and D1 appears on the left in B). A failed check returns its visual evidence to the Painter for regeneration (the feedback loop in Fig.~\ref{fig:framework}, with at most three attempts per step). Once verified, $I_t^\star$ is locked in and passed to the Director.

\subsection{Director: Geometry-Aware Process Rendering}
Given the current real frame $I_t$ and the verified frame $I_t^\star$, the \textbf{Director} synthesizes the transition in between. For interactions within a single space, the Director writes a motion description from $I_t$, $I_t^\star$, and the action text, and renders the action in one fixed camera shot. For moves between spaces, it reads from $L$ the camera orientation difference $\delta = (\phi_d - \phi_s) \bmod 360^{\circ}$ between source $r_s$ and destination $r_d$, and combines it with the portal geometry to specify the on-screen exit and entry sides across two fixed shots joined by a cut. For instance, when adjacent spaces share the same camera heading ($\delta=0^{\circ}$ in Fig.~\ref{fig:framework}), a character exiting through a doorway on the right naturally enters the next space from the left. Together with $I_t$ and $I_t^\star$, these instructions condition the FL2V video model to render a 10-second transition clip $V_t$. Once $V_t$ is rendered, its actual last frame $I_{t+1} = \mathrm{Tail}(V_t)$ both seeds the next turn and updates the space memory in $\mathcal{A}_{t+1}$ as the latest visual observation of the visited space (the feedback arrow to the next-turn panel in Fig.~\ref{fig:framework}), preserving scene and object changes across subsequent turns and revisits.

\section{Benchmark}
Unlike existing video benchmarks that separately evaluate clip-level visual quality~\cite{vbench}, prescribed multi-turn instruction adherence~\cite{wbench}, disappear-and-reappear object recovery under occlusion~\cite{memobench}, or paired counterfactual interventions~\cite{whatifworld,twinrollouts}, we build an interactive benchmark that jointly evaluates multi-space execution consistency and pre-render branching admissibility toward a shared goal.

\subsection{Evaluation Suites}
Built on 40 natural long-horizon tasks authored with Hy4~\cite{hy4} across diverse indoor, outdoor, and mixed scenes (10 each across 1--4 connected spaces), the benchmark comprises two suites:
(1) \textbf{Multi-Step Interactive Execution} evaluates rollout consistency over these 40 multi-turn episodes.
(2) \textbf{Branching Admissibility} evaluates pre-render proposal gating at their opening decision points across 571 truth-verified, player-phrased proposals: 237 valid routes and 334 disguised infeasible counterfactuals involving nonexistent places ($n=80$), nonexistent objects ($n=80$), or unsatisfied preconditions ($n=174$).

\subsection{Evaluation Metrics}
For \textbf{Multi-Step Interactive Execution}, we evaluate each step across \textbf{Action}, \textbf{World State}, and \textbf{Spatial continuity}. Automated Hy4 evaluation scores all ten items, while blind human raters annotate four core metrics: \emph{Action} (action completion), \emph{State} (object holding and support), \emph{Entry} (destination and entry side), and \emph{Exit} (exit doorway direction). Steps where lost camera framing hides the target are flagged separately as \emph{Unclear} (Sec.~\ref{sec:exp_interface}); the full ten-item list and scoring rubric are detailed in the supplementary material (Supplementary Table~S1).

For \textbf{Branching Admissibility}, we primarily evaluate under an \textbf{Open} mode (\emph{admissible} / \emph{inadmissible} / \emph{unsure}, where \emph{unsure} counts as a failure), reporting the acceptance rate on valid proposals (\emph{Accept}), rejection rate on infeasible ones (\emph{Reject}), their mean (\emph{Balanced}), and the \emph{Unsure} rate. We also include a \textbf{Forced} binary control that disallows abstention to check whether high rejection relies on trivially refusing valid proposals.

\section{Experiments}
\subsection{Experimental Setup}
\paragraph{Models and backbones.} All high-level reasoning roles (Architect, Interpreter, Planner, Inspector, and the task authoring model) are served by the multimodal language model Hy4~\cite{hy4}. For generative rendering, we cross-pair two image generators (Qwen-Image-2.1~\cite{qwenimage21} and FLUX.2~[klein]~4B~\cite{flux2klein}) with two frozen FL2V video backbones (MiniMax-H3~\cite{minimaxh3} and LTX-2.5~\cite{ltx25}) to synthesize 10-second transition clips and compare Baseline against \method{} across combinations. Blind human evaluation and the reference-to-video (Ref2V) interface comparison in Sec.~\ref{sec:exp_interface} are conducted on Qwen-Image-2.1 + MiniMax-H3.

\paragraph{Ablation settings.}
To isolate each explicit representation, we evaluate four cumulative settings under a shared cross-space room-memory policy (initial room render on first entry, own last video frame on revisits):
\emph{Baseline} (Direct rollout) receives only the current frame and a user instruction;
\emph{+Layout} adds the spatial layout $L$ and camera transition rules;
\emph{+Assets} further adds entity identity crops and stable appearance descriptions from $\mathcal{A}_t$;
and \emph{\method{}} (Full framework) adds the scene graph $G_t$ with symbolic state deduction, deterministic state-change checklists, and target-frame checks.
Without $G_t$, the first three settings bypass symbolic planning and synthesize frames directly from the action instruction and their allowed layout or asset inputs.
All compared settings share identical frozen backbones, action sequences, initial frames, random seeds, and per-step compute budgets (at most three target-frame attempts and one video generation), rolling out autoregressively from their own generated video tails to isolate the contribution of explicit world modeling.

\begin{table}[t]
\centering\small
\setlength{\tabcolsep}{1.8pt}
\caption{Automated Hy4 evaluation across image and video backbone combinations (FLUX.2+H3 evaluates Baseline vs.\ \method{}; 10-item breakdown in Supplementary Table~S3).}
\label{tab:backbones}
\begin{tabular}{@{}lllcccc@{}}
\toprule
Image & Video & Setting & Act. & World & Spat. & All \\
\midrule
\multirow{4}{*}{Qwen2.1~\cite{qwenimage21}} & \multirow{4}{*}{H3~\cite{minimaxh3}}
& Baseline & .630 & .760 & .807 & .748 \\
& & +Layout & .647 & .734 & .876 & .777 \\
& & +Assets & .670 & .780 & .864 & .791 \\
& & \method{} & \textbf{.767} & \textbf{.878} & \textbf{.877} & \textbf{.862} \\
\midrule
\multirow{2}{*}{FLUX.2~\cite{flux2klein}} & \multirow{2}{*}{H3~\cite{minimaxh3}}
& Baseline & .517 & .547 & .823 & .636 \\
& & \method{} & \textbf{.600} & \textbf{.596} & \textbf{.889} & \textbf{.691} \\
\midrule
\multirow{4}{*}{Qwen2.1~\cite{qwenimage21}} & \multirow{4}{*}{LTX-2.5~\cite{ltx25}}
& Baseline & .586 & .677 & .698 & .675 \\
& & +Layout & .586 & .679 & \textbf{.746} & .699 \\
& & +Assets & .657 & .747 & .698 & .731 \\
& & \method{} & \textbf{.662} & \textbf{.827} & .712 & \textbf{.766} \\
\midrule
\multirow{4}{*}{FLUX.2~\cite{flux2klein}} & \multirow{4}{*}{LTX-2.5~\cite{ltx25}}
& Baseline & .471 & .502 & \textbf{.751} & .586 \\
& & +Layout & .557 & .541 & .679 & .607 \\
& & +Assets & .508 & .535 & .642 & .576 \\
& & \method{} & \textbf{.602} & \textbf{.687} & .686 & \textbf{.675} \\
\bottomrule
\end{tabular}
\end{table}

\begin{figure}[t]
\centering
\includegraphics[width=\linewidth]{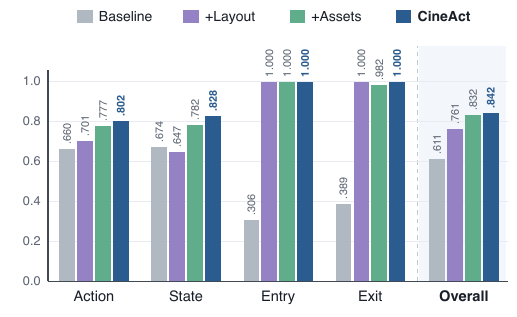}
\caption{\textbf{Blind human evaluation by 10 raters} over 10 sampled stories using Qwen-Image-2.1 + MiniMax-H3 (FL2V).}
\label{fig:main_human}
\end{figure}

\begin{figure*}[t]
\centering
\includegraphics[width=0.85\textwidth]{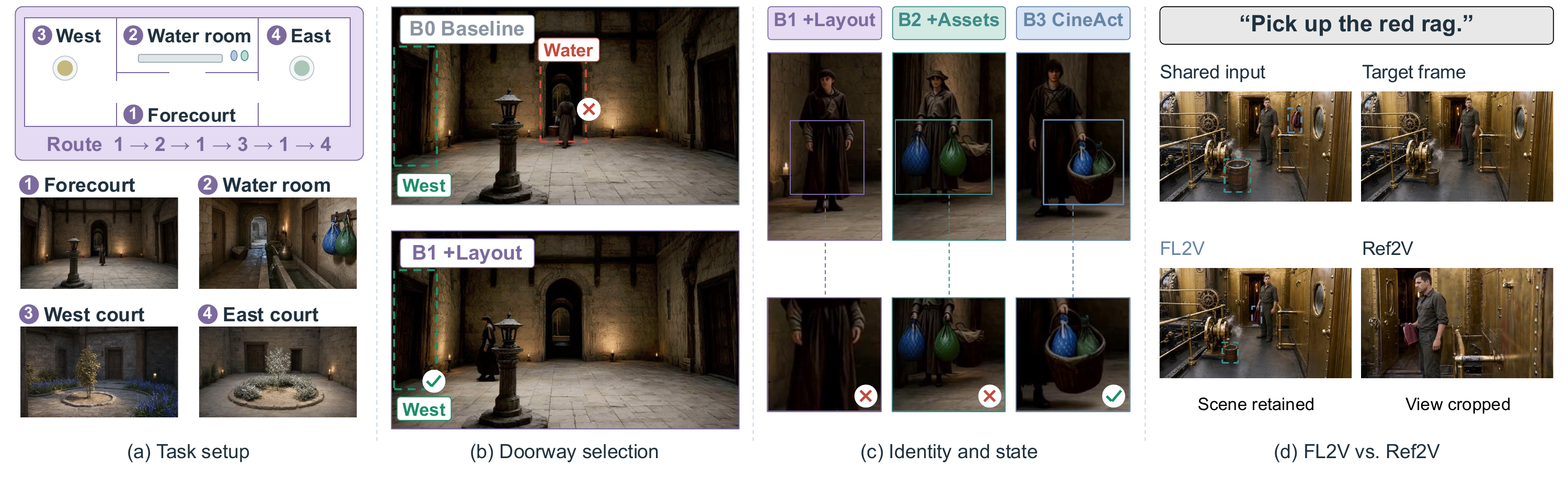}
\caption{\textbf{Qualitative comparisons.} \textbf{(a--c)} Multi-step execution: layout enables correct doorway routing (b), while full \method{} preserves identity and containment (c). \textbf{(d)} FL2V vs.\ Ref2V: verified \emph{Target frame} retains the room view, whereas \emph{Ref2V} crops the scene.}
\label{fig:qualitative}
\end{figure*}

\subsection{Multi-Step Execution}
\label{sec:exp_main}
Table~\ref{tab:backbones} and Fig.~\ref{fig:main_human} summarize the stepwise ablations on Multi-Step Interactive Execution across all four backbone combinations (Hy4 ratings; full 10-item breakdowns in Supplementary Table~S3) and blind human ratings by 10 raters on Qwen-Image-2.1 + MiniMax-H3:
\begin{itemize}
\item \textbf{Layout and cameras (+Layout vs.\ Baseline):} Adding spatial layout $L$ and camera transition rules improves spatial and overall consistency across backbones, effectively resolving cross-space disorientation.
\item \textbf{Asset library (+Assets vs.\ +Layout):} Introducing refreshed scene and instance references from $\mathcal{A}_t$ further boosts Action and World State under Qwen-Image-2.1, whereas lightweight FLUX.2~[klein]~4B~\cite{flux2klein} exhibits slight multi-reference editing variance before full \method{} recovers strongly.
\item \textbf{Scene graph and verification (\method{} vs.\ Baseline):} Adding $G_t$, transition specifications, and target-frame checks achieves the highest consistency across all four backbone pairings, lifting Action, World State, and overall scores over Baseline (.862 overall on Qwen2.1 + H3, +11.4 points) and demonstrating strong cross-backbone generalization.
\item \textbf{Blind human evaluation (Fig.~\ref{fig:main_human}):} Ten raters blindly scored 10 sampled stories on Qwen-Image-2.1 + MiniMax-H3 (40 anonymized story--setting units, 196 clips, 784 item ratings), corroborating the monotonic gain from .611 to .842 overall (+23.1 points) with best scores on every metric (Action .802, State .828, Entry/Exit 1.000).
\end{itemize}

Fig.~\ref{fig:qualitative}(a--c) illustrates this stepwise progression on a four-room watering task: given the setup and route in (a), \emph{Baseline} enters the wrong central arch toward the water room whereas \emph{+Layout} takes the correct \emph{West} door (b); after multiple room transitions, \emph{+Layout} drops the items and \emph{+Assets} holds both skins loose outside the basket, while \method{} keeps both packed inside the basket (c).

\begin{table}[t]
\centering\small
\setlength{\tabcolsep}{2.8pt}
\caption{Branching admissibility across 40 stories (237 valid + 334 invalid = 571 proposals). \emph{Open} counts \emph{Unsure} as failure; \emph{Forced} disallows abstention.}
\label{tab:branch}
\begin{tabular}{@{}llcccc@{}}
\toprule
Mode & Setting & Accept $\uparrow$ & Reject $\uparrow$ & Balanced $\uparrow$ & Unsure $\downarrow$ \\
\midrule
\multirow{4}{*}{Open}
& Baseline & .080 & .213 & .146 & 80.0\% \\
& +Layout & .089 & .287 & .188 & 78.1\% \\
& +Assets & .498 & .722 & .610 & 29.8\% \\
& \method{} & \textbf{.810} & \textbf{.964} & \textbf{.887} & \textbf{4.0\%} \\
\midrule
\multirow{4}{*}{Forced}
& Baseline & .262 & .970 & .616 & -- \\
& +Layout & .650 & .895 & .772 & -- \\
& +Assets & .738 & .934 & .836 & -- \\
& \method{} & \textbf{.827} & \textbf{.982} & \textbf{.905} & -- \\
\bottomrule
\end{tabular}
\end{table}

\begin{table}[t]
\centering\small
\setlength{\tabcolsep}{1.8pt}
\caption{Infeasible-proposal miss rate (not rejected $\downarrow$) under \emph{Open} mode. Best per row is bolded; underlined values mark layer-specific drops.}
\label{tab:branch_types}
\begin{tabular}{@{}lcccc@{}}
\toprule
Infeasible Category & Base. & +Layout & +Assets & \method{} \\
\midrule
Nonexistent Place ($n=80$) & .95 & \underline{.24} & .05 & \textbf{.00} \\
Nonexistent Object ($n=80$) & .80 & .91 & \underline{.30} & \textbf{.06} \\
Broken Precondition ($n=174$) & .71 & .84 & .37 & \textbf{.04} \\
\bottomrule
\end{tabular}
\end{table}

\subsection{Branching Admissibility}
\label{sec:exp_branch}
Interactive video generation requires more than following a single fixed script: given a shared goal, a system must accommodate diverse player-proposed branches by admitting valid alternative routes while filtering out requests that violate world constraints. Across the opening decision points of all 40 stories, we evaluate 571 player-phrased proposals authored with Hy4 (237 valid goal-reaching approaches and 334 plausible-looking infeasible counterfactuals). Each ablation setting judges proposal admissibility solely from the opening frame, task goal, and its allowed prior information ($L$, $\mathcal{A}_0$ containing initial views of all spaces and entity crops, or $G_0$) without generating video.

As shown in Tables~\ref{tab:branch} and~\ref{tab:branch_types}, pixel-only Baseline abstains (\emph{Unsure}) on 80.0\% of proposals under the \emph{Open} mode (especially in multi-space scenes) and accepts only .080 of valid branches (.146 balanced accuracy). Even under the \emph{Forced} control, Baseline achieves .970 rejection merely by refusing nearly all inputs, blocking 73.8\% of valid routes (.262 acceptance). Across the three counterfactual categories (Table~\ref{tab:branch_types}, where miss rate counts infeasible proposals answered either \emph{admissible} or \emph{unsure}), adding explicit representations progressively resolves each error type: \emph{+Layout} substantially reduces misses on \emph{Nonexistent Place}; \emph{+Assets} sharply cuts \emph{Nonexistent Object} misses and lowers overall abstention to 29.8\%; and adding $G_0$ (\method{}) further drives down \emph{Broken Precondition} misses, reducing abstention to 4.0\% while achieving high acceptance and rejection across both Open (.887 balanced) and Forced (.905 balanced) modes. To evaluate paired execution, raters blindly score 13 multi-route stories (26 trajectories/method across 1--4 spaces; Table~\ref{tab:paired_branch}) on per-route and both-route goal completion, route-specific consequence preservation, and zero cross-leak or rollback (jointly counted as \emph{both succeed \& distinct}). From identical initial states, \method{} achieves 1.000 both-route goal completion (vs.\ .615) and .923 \emph{both succeed \& distinct} (.900 on 2--4 spaces, $n{=}10$, vs.\ .500).

\begin{table}[t]
\centering\small
\setlength{\tabcolsep}{2.2pt}
\caption{Blind human evaluation of paired branching execution across 13 multi-route stories (26 trajectories per method).}
\label{tab:paired_branch}
\begin{tabular}{@{}lccc@{}}
\toprule
Metric & Baseline & \method{} & $\Delta$ \\
\midrule
Per-route goal completion & .769 & \textbf{1.000} & +.231 \\
Both-route goal completion & .615 & \textbf{1.000} & +.385 \\
Route-specific consequence & .769 & \textbf{1.000} & +.231 \\
No cross-leak or rollback & .846 & \textbf{.962} & +.115 \\
\midrule
Both succeed \& distinct ($n{=}13$) & .615 & \textbf{.923} & +.308 \\
Both succeed \& distinct (2--4 sp.) & .500 & \textbf{.900} & +.400 \\
\bottomrule
\end{tabular}
\end{table}

\subsection{Interface Analysis: FL2V vs.\ Ref2V}
\label{sec:exp_interface}
We compare our FL2V pipeline against Reference-to-Video (Ref2V), which conditions MiniMax-H3 directly on $I_t$ and asset crops over the same 10 stories. On visually determinable items, Ref2V achieves comparable Baseline-to-\method{} gains under the 10-item Hy4 evaluation (.713$\rightarrow$.875 vs.\ .733$\rightarrow$.856 for FL2V). However, Ref2V leaves 26.0\% (51/196) of object-holding/support (\emph{State}) steps visually \emph{Unclear} under human evaluation (vs.\ 0.0\% for FL2V), which is unacceptable for multi-turn interaction. Specifically, as shown in Fig.~\ref{fig:qualitative}(d), starting from the \emph{Shared input}, FL2V anchors the room framing via the verified \emph{Target frame}, so its output keeps both the held rag and the bucket and machinery in view (\emph{Scene retained}); by contrast, \emph{Ref2V} drifts into a close-up that crops out the bucket and surrounding room (\emph{View cropped}), preventing users from inspecting or selecting them in subsequent turns (Supplementary Table~S4).

\section{Limitations}
While \method{} improves multi-turn consistency and branching reliability, three trade-offs remain. First, skipping per-shot 3D reconstruction and fine-tuning keeps the pipeline lightweight and consistent, but limits each room to a fixed camera with cut transitions rather than continuous 3D camera motion. Second, sequential deduction, iterative verification, and 10-second FL2V synthesis yield turn-based rather than real-time streaming latency. Third, rooms and entities are fixed up front by Architect; adding new rooms on the fly, post-clip state re-sync, and streaming backbones is left to future work.

\section{Conclusion}
We formulated goal-grounded interactive video generation and presented \method{}, a state-first agentic framework that decouples high-level world-state reasoning from low-level pixel rendering. By maintaining explicit spatial layouts with cameras, dynamically refreshed asset libraries, and scene graphs, and by locking in post-action endpoints through pre-rendering target-frame verification, \method{} enables reliable multi-turn execution and pre-render gating of open-ended branching proposals. Experiments across multi-step execution, counterfactual branching admissibility, and interface comparisons validate each explicit representation's specialized role.

{\small\bibliographystyle{ieeenat_fullname}\bibliography{references}}

\clearpage
\setcounter{table}{0}
\setcounter{figure}{0}
\setcounter{section}{0}
\renewcommand{\thetable}{S\arabic{table}}
\renewcommand{\thefigure}{S\arabic{figure}}
\renewcommand{\thesection}{S\arabic{section}}
\renewcommand{\theHtable}{S\arabic{table}}
\renewcommand{\theHfigure}{S\arabic{figure}}
\maketitlesupplementary

\subsection*{Backbone Specifications}
All reasoning agents and automated evaluators run on the multimodal language model Hy4~\cite{hy4}. For image editing, we use Qwen-Image-2.1~\cite{qwenimage21} and the distilled FLUX.2~[klein]~4B~\cite{flux2klein}. For video synthesis, both the FL2V and Ref2V interfaces of MiniMax-H3~\cite{minimaxh3} use the 8-step distilled variants, while LTX-2.5~\cite{ltx25} runs in full BF16 non-distilled two-stage mode (30 steps).

\subsection*{System Representations and Execution Walkthrough}
Here we detail the three representations from Sec.~4 and summarize one execution turn in Algorithm~1. Fig.~\ref{fig:supp_case_seedlings} shows a concrete example on the three-terrace seedling task: panels~(a)--(d) give the layout $L$, initial views, scene graph $G_1$, and asset crops $\mathcal{A}_1$, while panels~(e)--(f) trace two rollouts. At each turn, the Planner checks preconditions on $G_t$, retrieves only the destination view and off-screen entity crops (Lines~4--5), verifies the target frame before rendering $V_t$ (Lines~7--12), and writes $\mathrm{Tail}(V_t)$ back only to the visited terrace (Line~13).

\begin{itemize}
\item \textbf{Spatial layout and cameras ($L$):} Spaces are non-overlapping 2D axis-aligned rectangles with portals on shared walls, each observed by one fixed camera ($65^\circ$ horizontal FOV, placed on a portal-free wall facing an exit). Geometric projection derives each doorway's on-screen side (\emph{left}, \emph{right}, or \emph{far wall}) for Painter/Director prompting and Inspector verification.
\item \textbf{Asset library ($\mathcal{A}_t$):} Combines per-space visual memory with static entity identity crops extracted at initialization. Only actual video tails $I_{t+1}=\mathrm{Tail}(V_t)$ (never intermediate target images $I_t^\star$) update the visited space's memory in $\mathcal{A}_{t+1}$; conditioning uses the destination space's latest memory plus crops for operated and off-screen entities, skipping actors already visible in $I_t$.
\item \textbf{Scene graph ($G_t$) and symbolic engine:} Nodes store location/ownership pointers (\emph{in}, \emph{on}, \emph{held}), portability, carrying capacity, and state attributes. Before rendering, the Planner checks preconditions and goal reachability via breadth-first search, advances the separate planned graph $G_{t+1}$, and deterministically diffs $(G_t, G_{t+1})$ into $\mathrm{Spec}_t$.
\end{itemize}

\begin{figure*}[!t]
\centering
\setlength{\tabcolsep}{4pt}
\begin{tabularx}{\textwidth}{@{}rlX@{}}
\toprule
\multicolumn{3}{@{}l@{}}{\textbf{Algorithm 1:} Single-turn execution of \method{} at turn $t$} \\
\midrule
\multicolumn{3}{@{}l@{}}{\textbf{Input:} Current frame $I_t$, layout $L$, assets $\mathcal{A}_t$, scene graph $G_t$, user input $u_t$} \\
1: & $a_t \leftarrow \mathrm{Interpreter}(u_t, I_t, \mathcal{A}_t, G_t)$ & // Ground input to entity and portal IDs \\
2: & \textbf{if} $\neg \mathrm{CheckPreconditions}(G_t, a_t)$ \textbf{or} $\neg \mathrm{GoalReachable}(G_t \oplus \Delta_t, g)$ \textbf{then} & \\
3: & \quad \textbf{return} \textbf{Reject}$(u_t)$ & // Pre-render gating before pixel synthesis \\
4: & $G_{t+1} \leftarrow G_t \oplus \Delta_t$; \quad $\mathrm{Spec}_t \leftarrow \mathrm{Diff}(G_t, G_{t+1}, L)$ & // Planned graph + deterministic diff \\
5: & $R_t \leftarrow \mathrm{RetrieveRefs}(\mathcal{A}_t, I_t, \mathrm{Spec}_t)$ & // Dest.\ space memory + non-visible crops \\
6: & $\mathrm{fb} \leftarrow \emptyset$ & \\
7: & \textbf{for} $\mathrm{attempt} = 1 \dots 3$ \textbf{do} & \\
8: & \quad $I_t^\star \leftarrow \mathrm{Painter}(I_t, R_t, \mathrm{Spec}_t, \mathrm{fb})$ & // Synthesize candidate target end frame \\
9: & \quad $(v, \mathrm{fb}) \leftarrow \mathrm{Inspector}(I_t^\star, I_t, R_t, \mathrm{Spec}_t, L)$ & // Semantic + doorway-side projection checks \\
10: & \quad \textbf{if} $v = \textbf{Pass}$ \textbf{then break} & \\
11: & $C_t \leftarrow \mathrm{CameraRule}(L, r_s \to r_d)$ & // 1 fixed shot (same) or 2 shots + cut (cross) \\
12: & $V_t \leftarrow \mathrm{Director}(I_t, I_t^\star, \mathrm{Spec}_t, C_t)$ & // Frozen FL2V video rendering \\
13: & $I_{t+1} \leftarrow \mathrm{Tail}(V_t)$; \quad $\mathcal{A}_{t+1} \leftarrow \mathrm{UpdateMemory}(\mathcal{A}_t, r_d, I_{t+1})$ & // Refresh visited space memory only \\
14: & \textbf{return} $(V_t, I_{t+1}, \mathcal{A}_{t+1}, G_{t+1})$ & \\
\bottomrule
\end{tabularx}
\end{figure*}

\subsection*{Evaluation Metrics and Human--Automated Agreement}
\begin{figure*}[!t]
\centering
\includegraphics[width=\textwidth,height=.90\textheight,keepaspectratio]{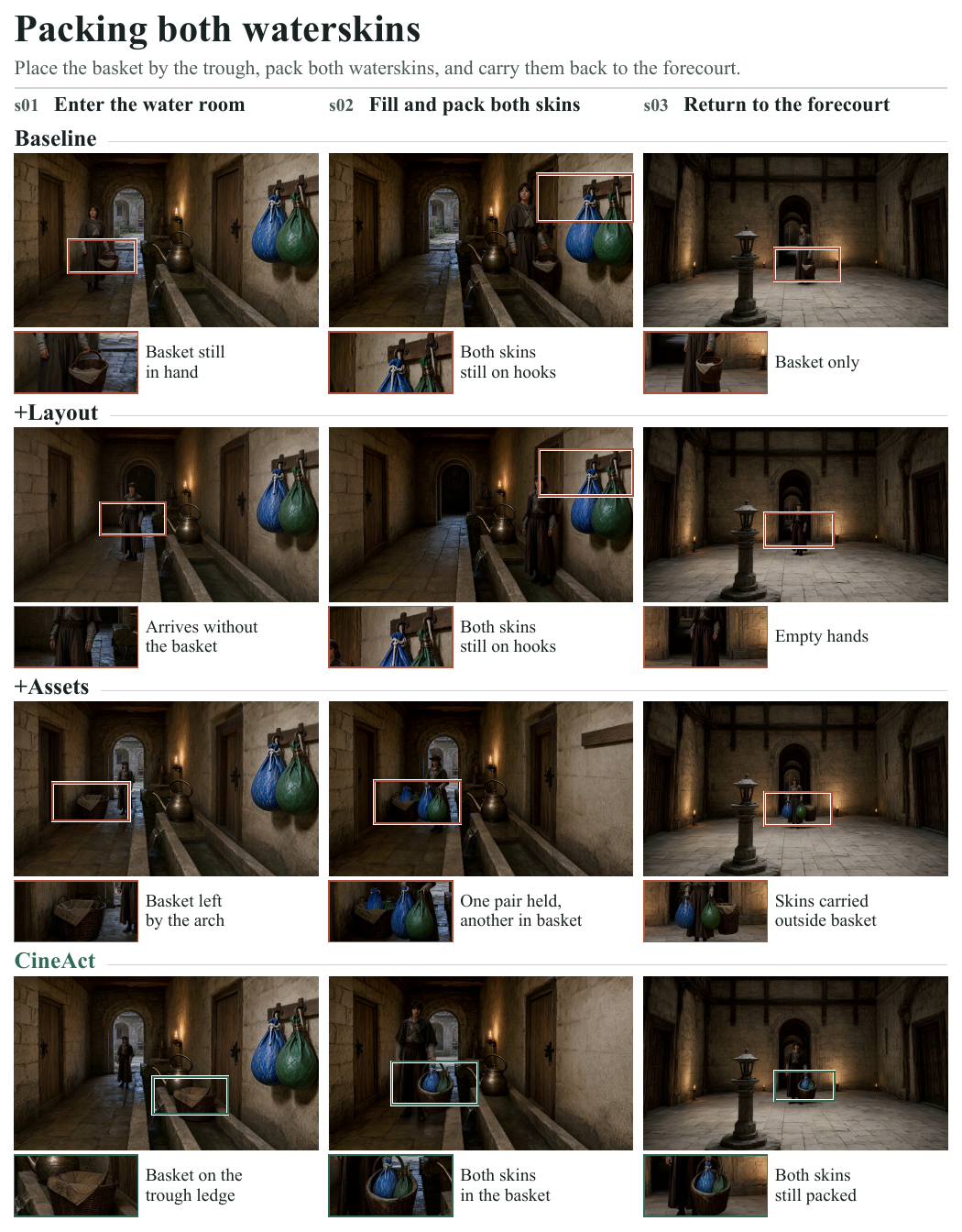}
\caption{\textbf{Waterskin pickup and transport} (courtyard watering, steps 1--3, Qwen-Image-2.1 and MiniMax-H3). Baseline and +Layout leave both waterskins on the hooks. +Assets duplicates the pair at pickup and carries the skins outside the basket on return. \method{} carries both in the basket. The insets enlarge the boxed regions of the video frames.}
\label{fig:supp_watering_p1}
\end{figure*}

\begin{figure*}[!t]
\centering
\includegraphics[width=\textwidth,height=.90\textheight,keepaspectratio]{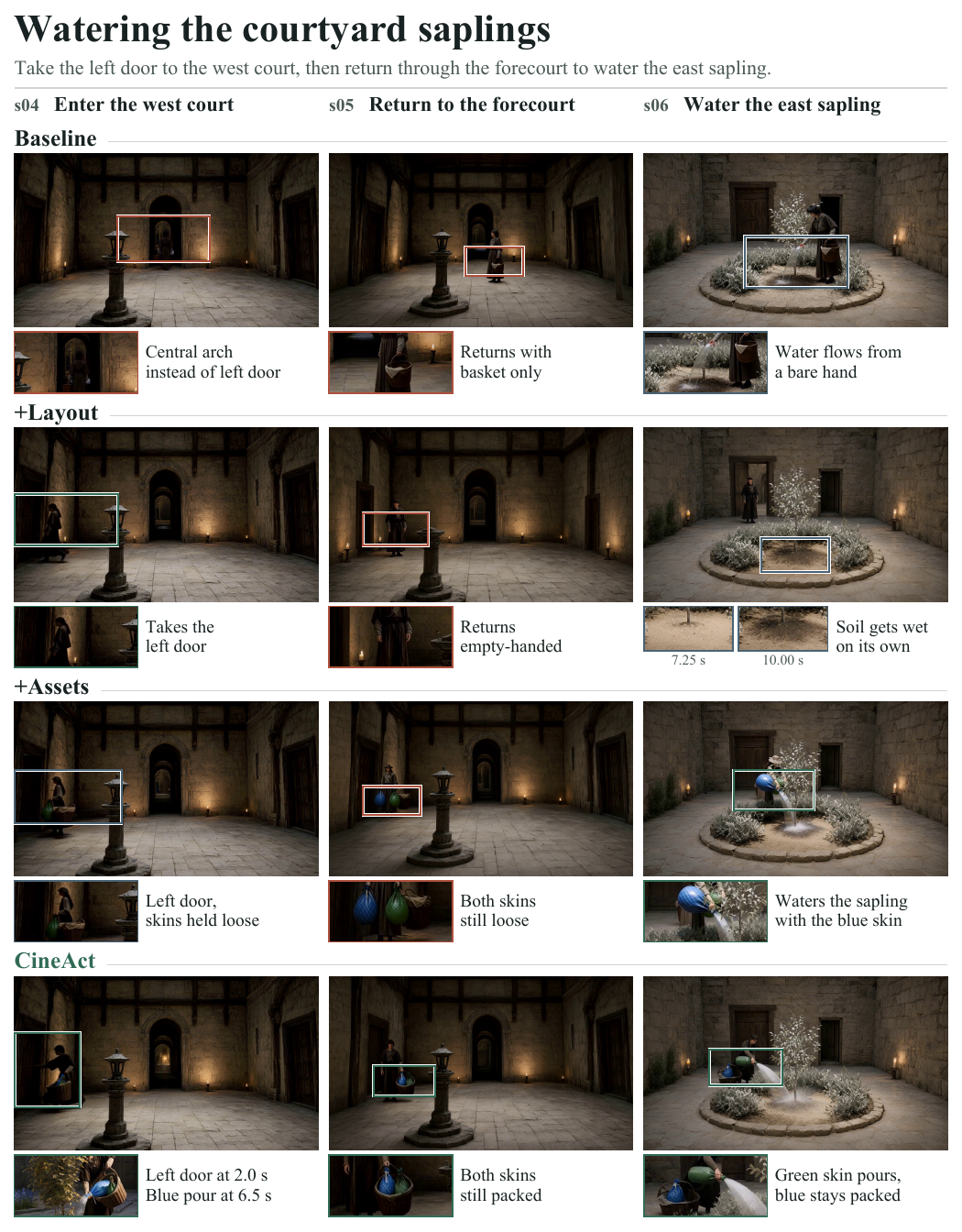}
\caption{\textbf{Door choice and watering} (courtyard watering, steps 4--6). Baseline takes the central arch on the way west. \method{} takes the left door and waters the west sapling with the blue skin (step-4 inset). In the east court, water flows from Baseline's bare hand. With +Layout, the soil around the sapling gets wet on its own between 7.25 and 10.00 seconds. +Assets waters the east sapling with the blue skin. \method{} uses the green skin and keeps the blue one in the basket. The cut in step 6 leaves \method{}'s final doorway crossing unconfirmed.}
\label{fig:supp_watering_p2}
\end{figure*}

Table~\ref{tab:metrics} lists all ten per-step questions used in Multi-Step Interactive Execution, and Figs.~\ref{fig:supp_watering_p1}--\ref{fig:supp_watering_p2} show all six steps of the courtyard watering task across the four ablation settings. On the four shared items (Action, State, Entry, Exit) for Qwen-Image-2.1 + MiniMax-H3, Hy4 matches blind human ratings 73.0\% of the time in binary agreement (64.4\% exact three-tier) and follows the same ablation order as Table~1 and Fig.~3 of the main paper. Table~\ref{tab:supp_human_backbones} extends this blind human study to two LTX-2.5 pairings on the same 10 stories. \method{} ranks first under all three pairings (.842, .748, and .614; +23.1, +18.0, and +17.6 points over Baseline), matching the automated cross-backbone results in Table~\ref{tab:supp_hy4_full}.

\begin{table}[h]
\centering\small
\setlength{\tabcolsep}{3.5pt}
\caption{Per-step evaluation items across the three consistency dimensions on Multi-Step Interactive Execution. All ten items are scored in automated Hy4 evaluation; $\star$ marks the four representative core items (Action, State, Entry, Exit) used for fast human annotation.}
\label{tab:metrics}
\begin{tabular}{@{}llp{4.5cm}@{}}
\toprule
ID & Dim. & Evaluation Question \\
\midrule
A1$^\star$ & Action & Was the requested action visibly executed and held? \\
\midrule
W1 & World State & Is the operated object the exact specified instance? \\
W2 & World State & Are entities free from duplication or vanishing? \\
W3$^\star$ & World State & Is the object held or supported on the surface correctly? \\
W4 & World State & Did the requested state attribute change as intended? \\
W5 & World State & Do removed or carried-away objects stay absent? \\
\midrule
S1$^\star$ & Spatial & Reaches destination space from correct entry side? \\
S2$^\star$ & Spatial & Departs source space toward correct doorway direction? \\
S3 & Spatial & Is camera cut placed on cross-space steps only? \\
S4 & Spatial & Do room layout and relative scale remain consistent? \\
\bottomrule
\end{tabular}
\end{table}

\begin{table}[h]
\centering\small
\setlength{\tabcolsep}{1.8pt}
\caption{Blind human evaluation (story-equal averages, three-tier 0/0.5/1 scale) across image and video backbone combinations on Multi-Step Interactive Execution.}
\label{tab:supp_human_backbones}
\begin{tabular}{@{}lllccccc@{}}
\toprule
Image & Video & Setting & Action & State & Entry & Exit & All \\
\midrule
\multirow{4}{*}{Qwen2.1} & \multirow{4}{*}{H3}
& Baseline & .660 & .674 & .306 & .389 & .611 \\
& & +Layout & .701 & .647 & \textbf{1.000} & \textbf{1.000} & .761 \\
& & +Assets & .777 & .782 & \textbf{1.000} & .982 & .832 \\
& & \method{} & \textbf{.802} & \textbf{.828} & \textbf{1.000} & \textbf{1.000} & \textbf{.842} \\
\midrule
\multirow{4}{*}{Qwen2.1} & \multirow{4}{*}{LTX-2.5}
& Baseline & .613 & .676 & .310 & .279 & .568 \\
& & +Layout & .569 & .668 & .626 & .524 & .627 \\
& & +Assets & .548 & .789 & .821 & \textbf{.914} & .711 \\
& & \method{} & \textbf{.619} & \textbf{.808} & \textbf{.907} & .871 & \textbf{.748} \\
\midrule
\multirow{4}{*}{FLUX.2} & \multirow{4}{*}{LTX-2.5}
& Baseline & .479 & .449 & .574 & .148 & .438 \\
& & +Layout & .497 & .395 & .724 & .638 & .506 \\
& & +Assets & .455 & .434 & .650 & \textbf{.736} & .503 \\
& & \method{} & \textbf{.565} & \textbf{.616} & \textbf{.829} & .679 & \textbf{.614} \\
\bottomrule
\end{tabular}
\end{table}

\begin{table*}[!t]
\centering\small
\setlength{\tabcolsep}{2.8pt}
\caption{Detailed 10-item Hy4 evaluation across all four image--video backbone combinations on Multi-Step Interactive Execution. For FLUX.2~[klein]~4B + MiniMax-H3, only Baseline and \method{} were evaluated.}
\label{tab:supp_hy4_full}
\begin{tabular}{@{}lll c ccccc c cccc c c@{}}
\toprule
& & & Action & \multicolumn{6}{c}{World State} & \multicolumn{5}{c}{Spatial} & \\
\cmidrule(lr){4-4}\cmidrule(lr){5-10}\cmidrule(lr){11-15}
Image & Video & Setting & A1$^\star$ & W1 & W2 & W3$^\star$ & W4 & W5 & Avg. & S1$^\star$ & S2$^\star$ & S3 & S4 & Avg. & Overall \\
\midrule
\multirow{4}{*}{Qwen2.1} & \multirow{4}{*}{H3}
& Baseline & .630 & .946 & .835 & .719 & .390 & .725 & .760 & .492 & .596 & .712 & .922 & .807 & .748 \\
& & +Layout & .647 & .973 & .850 & .637 & .460 & .688 & .734 & \textbf{.932} & \textbf{.990} & .745 & \textbf{.931} & .876 & .777 \\
& & +Assets & .670 & .970 & .825 & .755 & .489 & .682 & .780 & .910 & .924 & .700 & .916 & .864 & .791 \\
& & \method{} & \textbf{.767} & \textbf{.993} & \textbf{.870} & \textbf{.879} & \textbf{.652} & \textbf{.919} & \textbf{.878} & .925 & .941 & \textbf{.800} & .904 & \textbf{.877} & \textbf{.862} \\
\midrule
\multirow{2}{*}{FLUX.2~\cite{flux2klein}} & \multirow{2}{*}{H3}
& Baseline & .517 & \textbf{.889} & \textbf{.619} & .473 & .429 & .362 & .547 & .649 & .736 & .711 & .925 & .823 & .636 \\
& & \method{} & \textbf{.600} & .833 & .503 & \textbf{.642} & \textbf{.537} & \textbf{.598} & \textbf{.596} & \textbf{.804} & \textbf{.875} & \textbf{.817} & \textbf{.930} & \textbf{.889} & \textbf{.691} \\
\midrule
\multirow{4}{*}{Qwen2.1} & \multirow{4}{*}{LTX-2.5~\cite{ltx25}}
& Baseline & .586 & .963 & .704 & .683 & .377 & .525 & .677 & .713 & .619 & .476 & \textbf{.935} & .698 & .675 \\
& & +Layout & .586 & .925 & \textbf{.827} & .539 & .533 & .537 & .679 & .662 & .883 & \textbf{.605} & .888 & \textbf{.746} & .699 \\
& & +Assets & .657 & .950 & .684 & .745 & \textbf{.644} & .733 & .747 & \textbf{.917} & \textbf{1.000} & .447 & .865 & .698 & .731 \\
& & \method{} & \textbf{.662} & \textbf{.971} & .813 & \textbf{.827} & .512 & \textbf{.925} & \textbf{.827} & .849 & .821 & .470 & .888 & .712 & \textbf{.766} \\
\midrule
\multirow{4}{*}{FLUX.2~\cite{flux2klein}} & \multirow{4}{*}{LTX-2.5~\cite{ltx25}}
& Baseline & .471 & \textbf{.979} & .568 & .415 & .315 & .400 & .502 & .457 & .762 & \textbf{.657} & \textbf{.936} & \textbf{.751} & .586 \\
& & +Layout & .557 & .852 & \textbf{.633} & .429 & .360 & .558 & .541 & \textbf{.768} & .929 & .453 & .848 & .679 & .607 \\
& & +Assets & .508 & .882 & .499 & .544 & .329 & .433 & .535 & .721 & .857 & .425 & .799 & .642 & .576 \\
& & \method{} & \textbf{.602} & .971 & .556 & \textbf{.695} & \textbf{.569} & \textbf{.725} & \textbf{.687} & .636 & \textbf{.964} & .420 & .855 & .686 & \textbf{.675} \\
\bottomrule
\end{tabular}
\end{table*}

\subsection*{Interface Comparison: Target-Frame FL2V vs.\ Ref2V}
Table~\ref{tab:fl2v_vs_ref2v} gives the full human and Hy4 numbers behind Fig.~4(d) of the main paper on the same 10 stories. When a step is visibly scoreable, both interfaces gain from our representations: under Hy4, Ref2V rises from .488/.621/.900/.713 in Baseline to .725/.866/.959/.875 in \method{} (Action/World/Spatial/Overall). The bottleneck is framing stability. Across the 49 \emph{State} steps per setting, \textbf{FL2V has 0/196 Unclear items (0.0\%), while Ref2V has 12, 11, 15, and 13 Unclear steps from Baseline to \method{} (51/196, or 26.0\%)}. Without a verified target frame to anchor the shot (Fig.~4(d)), Ref2V zooms in and crops out the bucket and left machinery during pickup, leaving users unable to inspect or select those objects in the next turn.

\begin{table}[h]
\centering\small
\setlength{\tabcolsep}{1.5pt}
\caption{Target-Frame FL2V vs.\ Ref2V on the same 10 stories. FL2V human matches Fig.~3 of the main paper; Ref2V and Hy4 report story-equal averages.}
\label{tab:fl2v_vs_ref2v}
\begin{tabular}{@{}llccccc@{}}
\toprule
& & \multicolumn{4}{c}{Determinable Score} & \textbf{State Unclear} \\
\cmidrule(lr){3-6}
Track & Interface & Base. & +Lay. & +Ass. & \method{} & \textbf{Rate (\%)} \\
\midrule
Human & FL2V (Ours) & .611 & .761 & .832 & .842 & \textbf{0.0\%} \\
Human & Ref2V & .538 & .624 & .781 & .870 & 26.0\% \\
\midrule
Hy4 & FL2V (Ours) & .733 & .785 & .739 & .856 & -- \\
Hy4 & Ref2V & .713 & .771 & .836 & .875 & -- \\
\bottomrule
\end{tabular}
\end{table}

\subsection*{Paired Branching Execution: Per-Task Scores and Rollout Examples}
\begin{figure*}[p]
\centering
\includegraphics[width=\textwidth,height=.90\textheight,keepaspectratio]{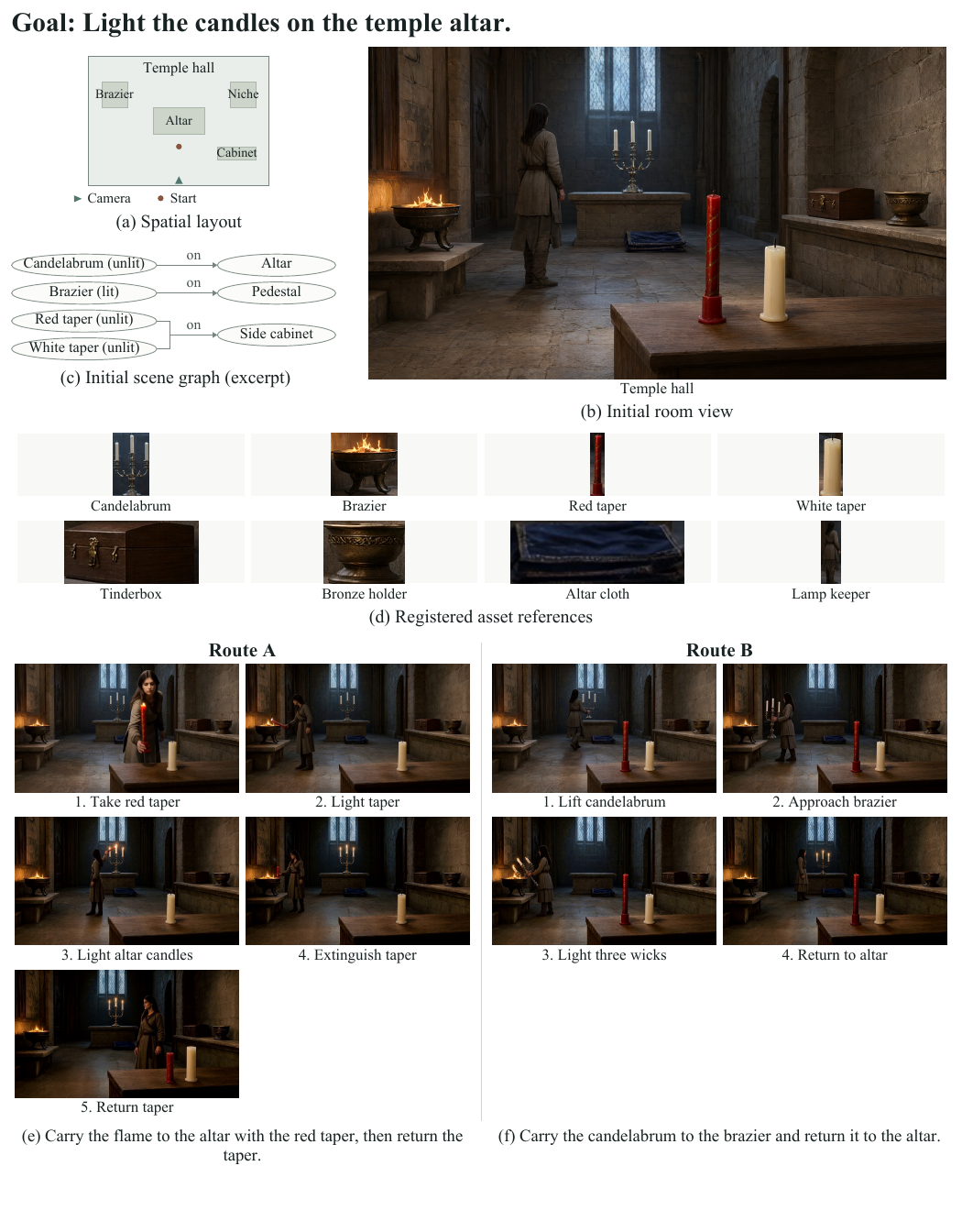}
\caption{\textbf{Lighting the altar candelabrum.} The shared opening contains an unlit candelabrum and a burning brazier. Route~A brings the flame to the altar with a red taper. Route~B brings the candelabrum to the flame. Each route shows one representative frame per turn, read left to right within each row.}
\label{fig:supp_case_altar}
\end{figure*}

\begin{figure*}[p]
\centering
\includegraphics[width=\textwidth,height=.90\textheight,keepaspectratio]{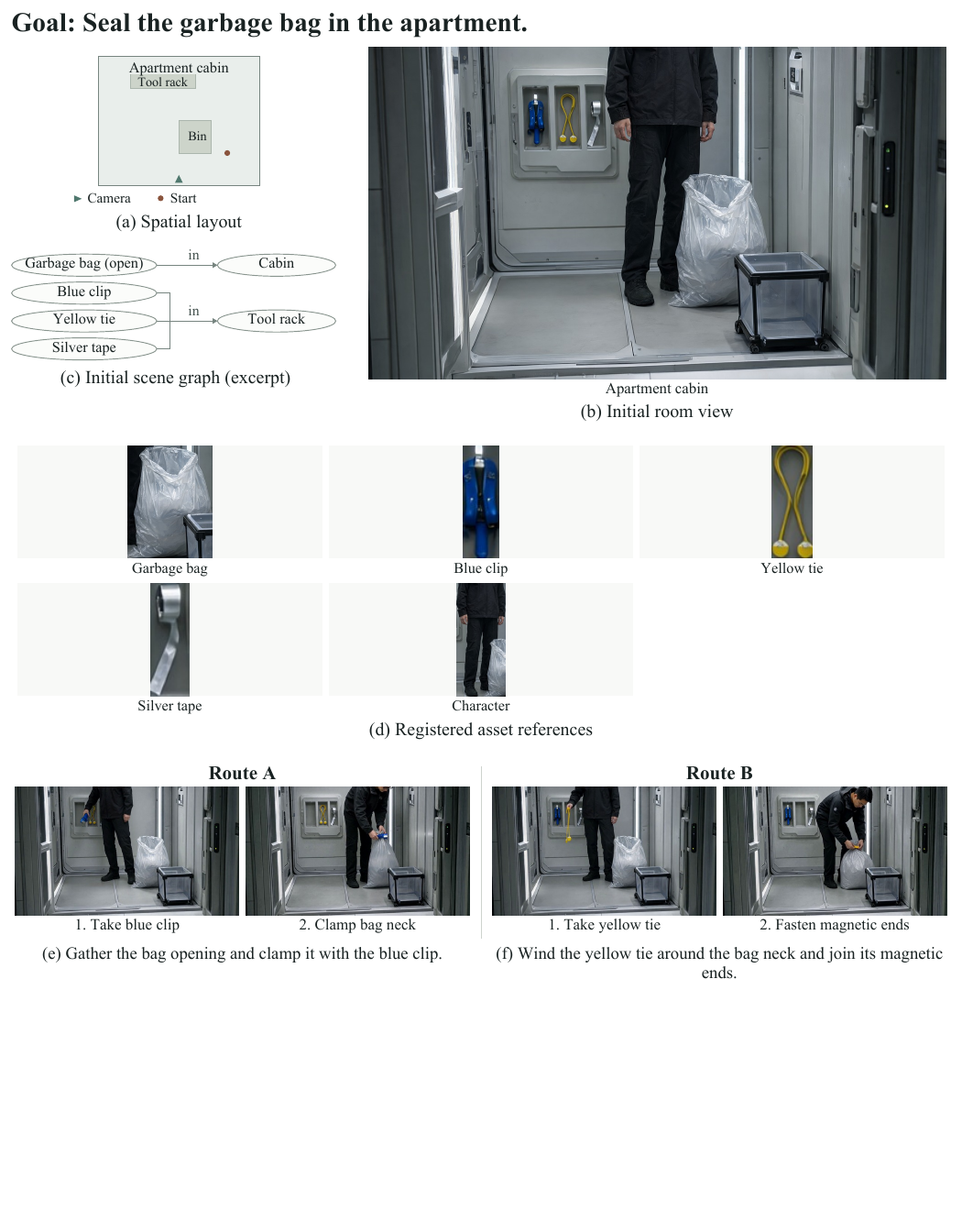}
\caption{\textbf{Sealing a bag in the apartment.} Both routes start with the same open bag and wall-mounted tool rack. They differ in the fastener selected for the bag neck. Each route shows one representative frame per turn, read left to right within each row.}
\label{fig:supp_case_bag}
\end{figure*}

\begin{figure*}[p]
\centering
\includegraphics[width=\textwidth,height=.90\textheight,keepaspectratio]{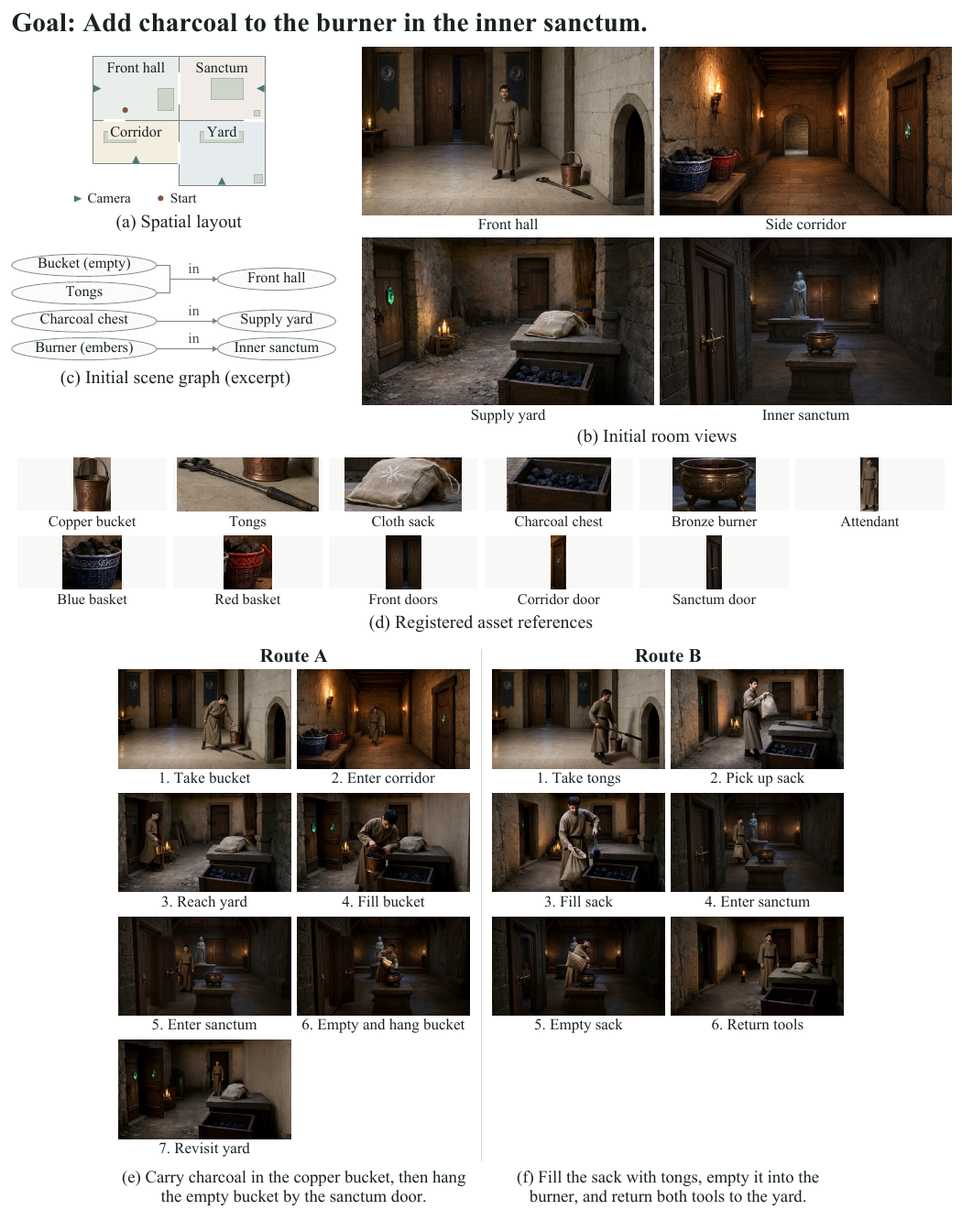}
\caption{\textbf{Delivering charcoal to the inner sanctum.} The routes draw charcoal from the same supply yard. The bucket route calls for leaving the empty bucket at the sanctum door. The sack route calls for returning the sack and tongs to the yard. Each route shows one representative frame per turn, read left to right within each row.}
\label{fig:supp_case_charcoal}
\end{figure*}

\begin{figure*}[p]
\centering
\includegraphics[width=\textwidth,height=.90\textheight,keepaspectratio]{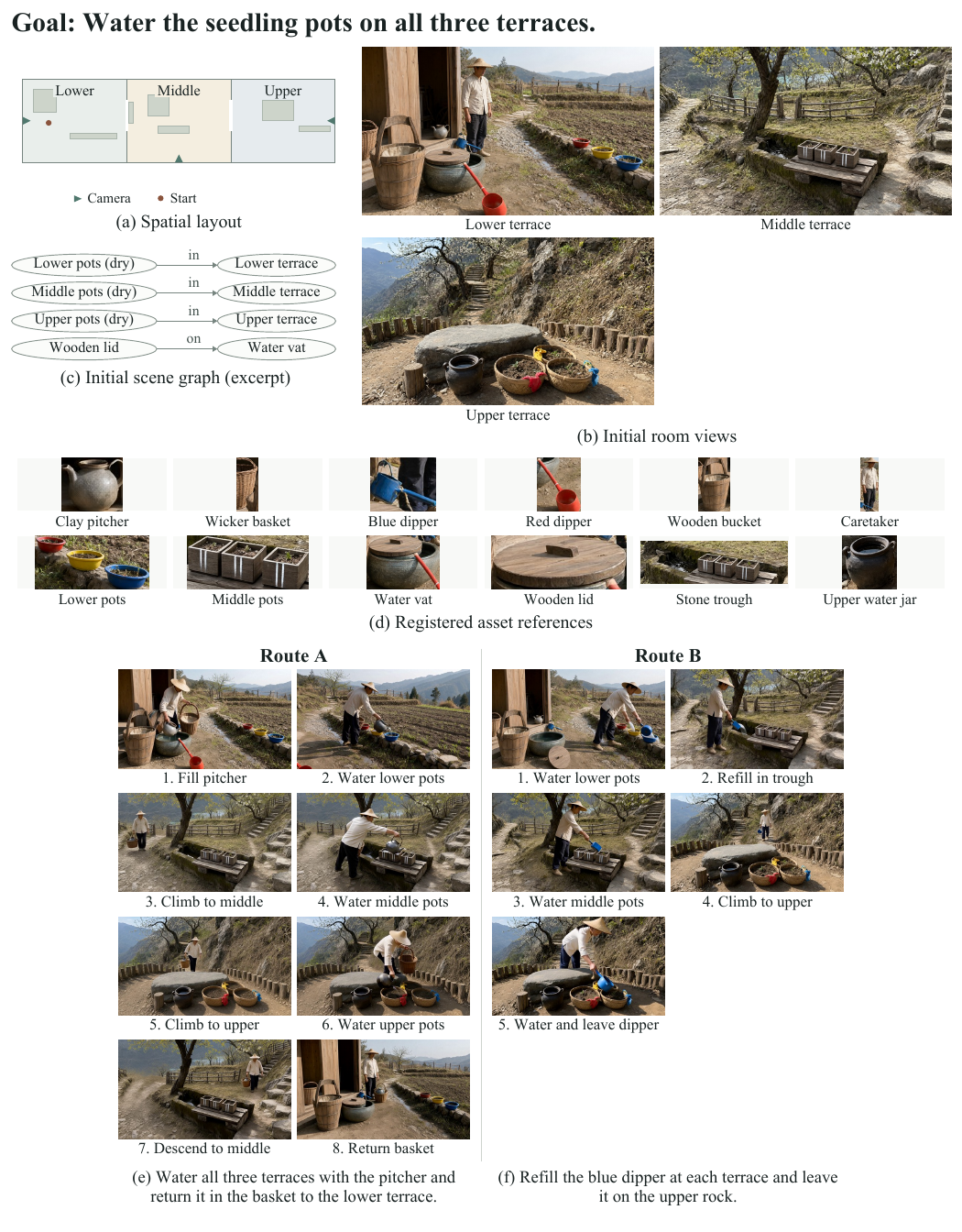}
\caption{\textbf{Watering seedling pots on three terraces.} The pitcher route ends back on the lower terrace with the pitcher and basket. The dipper route ends at the upper terrace, where the blue dipper is to be left on the rock. Each route shows one representative frame per turn, read left to right within each row.}
\label{fig:supp_case_seedlings}
\end{figure*}

\begin{figure*}[p]
\centering
\includegraphics[width=\textwidth,height=.90\textheight,keepaspectratio]{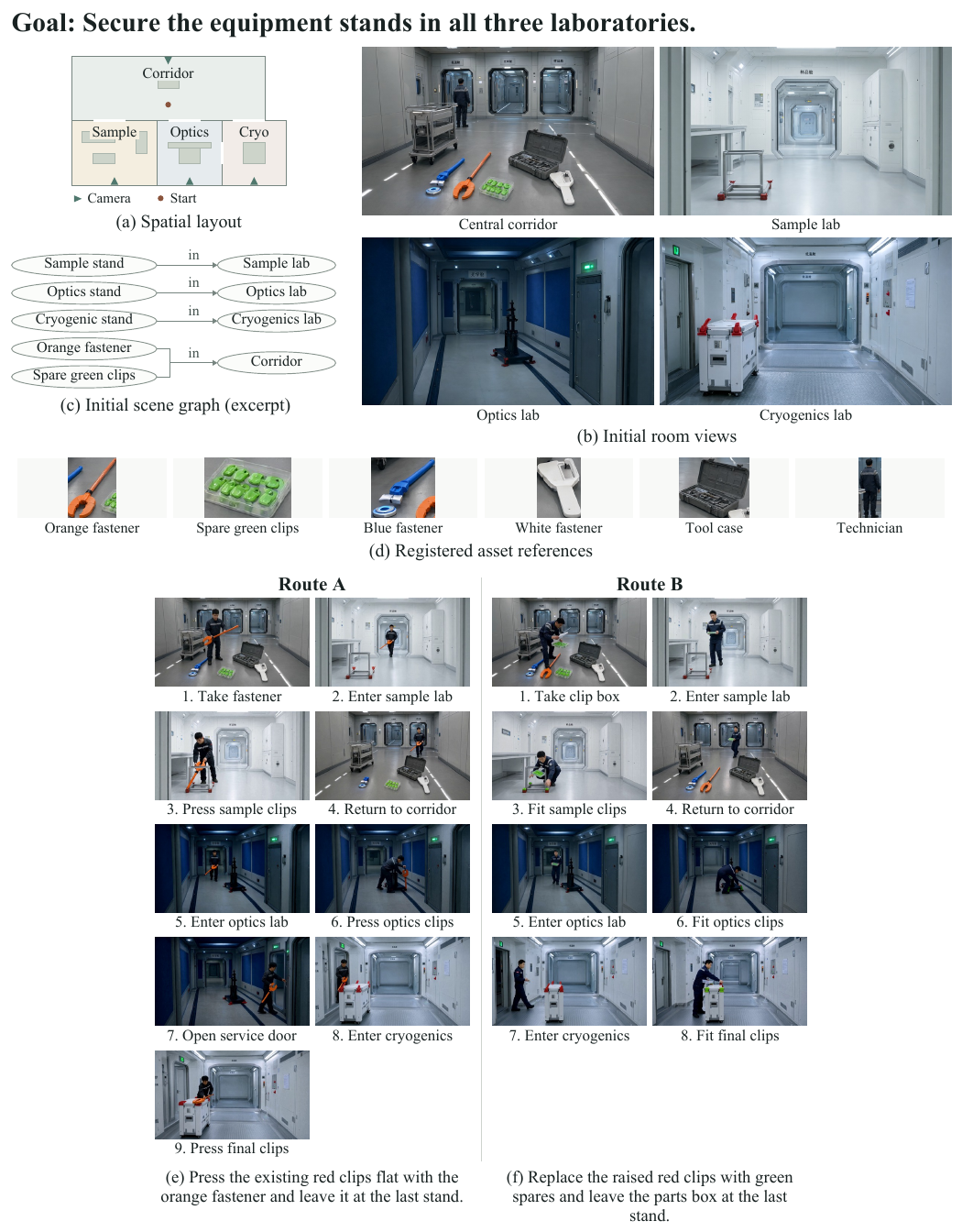}
\caption{\textbf{Securing stands in three laboratories.} Each stand starts with four raised red clips. One route uses the orange fastener to press down the existing clips. The other replaces them with green spares. Each route shows one representative frame per turn, read left to right within each row.}
\label{fig:supp_case_stands}
\end{figure*}

Table~\ref{tab:supp_paired_branch} breaks down the 13 paired branching tasks from Table~4 of the main paper (52 rollouts total, two routes per task) using the same 10-item per-step Hy4 protocol as Table~1 and Table~\ref{tab:supp_hy4_full}. Averaged equally across all 13 tasks, \method{} scores .884 on Route~1 (Action .802, World .903, Spatial .889 vs.\ .765 for Baseline) and .871 on Route~2 (.770/.896/.872 vs.\ .847 for Baseline). On the 10 multi-space tasks (2--4 spaces), \method{} reaches .878 on Route~1 and .838 on Route~2 (vs.\ .740 and .820 for Baseline). Figs.~\ref{fig:supp_case_altar}--\ref{fig:supp_case_stands} show five examples spanning 1--4 spaces: lighting the altar, sealing the bag, delivering charcoal, watering the terraces, and securing the lab stands.

\begin{table*}[!t]
\centering\footnotesize
\setlength{\tabcolsep}{2.1pt}
\caption{Per-task 10-item Hy4 scores for paired branching execution across 13 tasks, reported separately for Route~1 and Route~2 under Baseline and \method{} (Act., World, Spat., All).}
\label{tab:supp_paired_branch}
\begin{tabular}{@{}l cccc cccc cccc cccc@{}}
\toprule
& \multicolumn{4}{c}{Baseline Route 1} & \multicolumn{4}{c}{Baseline Route 2} & \multicolumn{4}{c}{\method{} Route 1} & \multicolumn{4}{c}{\method{} Route 2} \\
\cmidrule(lr){2-5}\cmidrule(lr){6-9}\cmidrule(lr){10-13}\cmidrule(lr){14-17}
Task & Act. & World & Spat. & All & Act. & World & Spat. & All & Act. & World & Spat. & All & Act. & World & Spat. & All \\
\midrule
Balcony flowers & .750 & .921 & 1.000 & \textbf{.919} & .833 & .962 & 1.000 & .955 & .625 & .861 & .875 & .833 & .833 & 1.000 & 1.000 & \textbf{.976} \\
Altar candles & .500 & .875 & .900 & .833 & .750 & .821 & 1.000 & .854 & .700 & .913 & .900 & \textbf{.882} & 1.000 & .929 & 1.000 & \textbf{.958} \\
Apartment bag & .750 & .722 & 1.000 & .800 & 1.000 & 1.000 & 1.000 & 1.000 & 1.000 & 1.000 & 1.000 & \textbf{1.000} & 1.000 & 1.000 & 1.000 & 1.000 \\
Library scrolls & .556 & .629 & .741 & .664 & .600 & .711 & .786 & .724 & .722 & .833 & .933 & \textbf{.861} & .600 & .684 & .821 & .724 \\
Beach buoy & .500 & .364 & .950 & .620 & 1.000 & 1.000 & .875 & \textbf{.957} & .750 & .821 & .900 & \textbf{.839} & 1.000 & .900 & .875 & .905 \\
Reef kelp & .571 & .654 & .667 & .647 & .875 & .923 & .625 & .820 & .786 & .911 & .850 & \textbf{.873} & 1.000 & 1.000 & .889 & \textbf{.964} \\
Sanctum charcoal & .643 & .778 & .857 & .791 & .500 & .727 & .861 & .756 & .857 & .920 & .905 & \textbf{.906} & .700 & .896 & .825 & \textbf{.847} \\
Cellar hook & .667 & .885 & .500 & \textbf{.750} & .500 & .875 & .875 & .821 & .667 & .773 & .625 & .705 & .500 & .844 & 1.000 & \textbf{.850} \\
Boiler coal & .800 & .861 & .833 & .843 & .750 & .967 & .917 & \textbf{.919} & .900 & .929 & .917 & \textbf{.921} & .750 & .833 & .833 & .823 \\
Stair blanket & .875 & .923 & .917 & .914 & .700 & .667 & .857 & .743 & 1.000 & 1.000 & .875 & \textbf{.953} & .600 & .925 & .821 & \textbf{.846} \\
Terrace seedlings & .625 & .560 & .609 & .589 & .600 & .809 & .808 & .782 & .875 & .950 & .913 & \textbf{.926} & .700 & .900 & .643 & .782 \\
Deck railings & .429 & .900 & .917 & .837 & .571 & .740 & .850 & .760 & .929 & .919 & 1.000 & \textbf{.941} & .571 & .904 & .842 & \textbf{.837} \\
Lab stands & .556 & .823 & .708 & .742 & .875 & .929 & .913 & \textbf{.915} & .611 & .914 & .865 & \textbf{.852} & .750 & .833 & .792 & .805 \\
\midrule
Mean (2--4 sp.) & .622 & .738 & .770 & .740 & .697 & .835 & .837 & .820 & .810 & .897 & .878 & \textbf{.878} & .717 & .872 & .834 & \textbf{.838} \\
Mean (All, n=13) & .632 & .761 & .815 & .765 & .735 & .856 & .874 & .847 & .802 & .903 & .889 & \textbf{.884} & .770 & .896 & .872 & \textbf{.871} \\
\bottomrule
\end{tabular}
\end{table*}

\subsection*{Branching Admissibility: Space Stratification and Counterfactual Breakdown}
Table~\ref{tab:supp_admissibility_spaces} splits the 40-story branching admissibility benchmark (Tables~2--3 of the main paper) by space count and counterfactual error type. We evaluate 10 stories per space count; filtering 56 ambiguous items (8.9\%) from 627 proposals leaves 571 verified queries. Under the open setting, pixel-only baselines answer \emph{Unsure} on 47.8\% of single-space queries and 87.1\%--95.1\% in 2--4 spaces, because off-screen rooms are absent from the current frame. \method{} keeps abstention at 2.7\%--5.1\% across all space counts. In the forced two-choice check, Baseline rejects 97.0\% of invalid proposals mainly by rejecting almost everything (including valid moves), while \method{} keeps both valid acceptance and invalid rejection high.

\begin{table*}[!t]
\centering\footnotesize
\setlength{\tabcolsep}{3.0pt}
\caption{Space-stratified branching admissibility (n=40 stories, 10 per layer, 571 kept queries) and unrejected counterfactual rate by category under the open rubric.}
\label{tab:supp_admissibility_spaces}
\begin{tabular}{@{}lcccc cccc cccc@{}}
\toprule
& \multicolumn{4}{c}{Query Counts} & \multicolumn{4}{c}{Open Abstain / Unsure Rate (\%)} & \multicolumn{4}{c}{Unrejected Invalid (\% of Cat.)} \\
\cmidrule(lr){2-5}\cmidrule(lr){6-9}\cmidrule(lr){10-13}
Layer / Category & Valid & Obj & Place & Precond & Base. & +Lay. & +Ass. & \method{} & Base. & +Lay. & +Ass. & \method{} \\
\midrule
1 space (n=10, tot=138) & 60 & 20 & 20 & 38 & 47.8 & 53.6 & 23.9 & \textbf{5.1} & -- & -- & -- & -- \\
2 spaces (n=10, tot=142) & 60 & 20 & 20 & 42 & 88.7 & 82.4 & 36.6 & \textbf{3.5} & -- & -- & -- & -- \\
3 spaces (n=10, tot=147) & 56 & 20 & 20 & 51 & 87.1 & 87.8 & 27.9 & \textbf{2.7} & -- & -- & -- & -- \\
4 spaces (n=10, tot=144) & 61 & 20 & 20 & 43 & 95.1 & 87.5 & 30.6 & \textbf{4.9} & -- & -- & -- & -- \\
\midrule
Nonexistent Object (n=80) & -- & 80 & -- & -- & -- & -- & -- & -- & 80.0 & 91.3 & 30.0 & \textbf{6.3} \\
Nonexistent Place (n=80) & -- & -- & 80 & -- & -- & -- & -- & -- & 95.0 & 23.8 & 5.0 & \textbf{0.0} \\
Broken Precond. (n=174) & -- & -- & -- & 174 & -- & -- & -- & -- & 70.7 & 83.9 & 37.4 & \textbf{4.0} \\
\midrule
Pooled Total (n=40, tot=571) & 237 & 80 & 80 & 174 & 80.0 & 78.1 & 29.8 & \textbf{4.0} & 78.7 & 71.3 & 27.8 & \textbf{3.6} \\
\bottomrule
\end{tabular}
\end{table*}

Below are concrete examples of the three counterfactual categories in Table~\ref{tab:supp_admissibility_spaces}. (1)~\emph{Nonexistent Place} asks to move into an unlinked or nonexistent room (e.g., going from the stone hall to an upper balcony absent from $L$ in courtyard watering); adding layout geometry drops unrejected errors from 95.0\% to 23.8\%. (2)~\emph{Nonexistent Object} names an item missing from $\mathcal{A}$ (e.g., wiping deck soot with a sponge instead of the red or blue rag); +Assets reduces these errors from 91.3\% to 30.0\%. (3)~\emph{Broken Precondition} violates capacity, location, or state rules (e.g., pouring into the stone basin before filling both waterskins), which requires \method{}'s symbolic state check (4.0\% vs.\ 37.4\% for +Assets).

\end{document}